\documentclass{bmvc2k_arxiv}
\usepackage{amsmath,amssymb}
\usepackage{rotating}
\usepackage[export]{adjustbox}
\usepackage[page]{appendix}
\usepackage{geometry}
\title{Deep Video Color Propagation}

\addauthor{Simone Meyer$^{\text{\scriptsize 1,}}$}{simone.meyer@inf.ethz.ch}{2}
\addauthor{Victor Cornill\`ere}{}{2}
\addauthor{Abdelaziz Djelouah}{aziz.djelouah@disneyresearch.com}{2}
\addauthor{Christopher Schroers}{}{2}
\addauthor{Markus Gross$^{\text{\scriptsize 1,}}$}{}{2}

\addinstitution{
Department of Computer Science\\ 
ETH Zurich\\
}
\addinstitution{
Disney Research
}

\runninghead{Meyer et al.}{Deep Video Color Propagation}

\DeclareMathOperator*{\argmin}{arg\,min}

\begin{document}

\maketitle

\begin{abstract}
Traditional approaches for color propagation in videos
rely on some form of matching between consecutive video frames. 
Using appearance descriptors, colors are then propagated both spatially and temporally.
These methods, however, are computationally expensive and do not take advantage of semantic information of the scene.
In this work we propose a deep learning framework for 
color propagation that combines a local strategy, to propagate
colors frame-by-frame ensuring temporal stability, and a global strategy,
using semantics for color propagation within a longer
range. Our evaluation shows the superiority of our strategy over existing
video and image color propagation methods as well 
as neural photo-realistic style transfer approaches. 
\end{abstract}

\section{Introduction}
Color propagation is an important problem in video processing
and has a wide range of applications. 
For example in movie making work-flow, 
where color modification for artistic purposes~\cite{Painting_with_pixels} 
plays an important role.
It is also used in the restoration 
and colorization of heritage 
footage~\cite{AmericaInColor} for more engaging experiences. 
Finally, the ability to faithfully propagate 
colors in videos can have a direct impact on video compression.

Traditional approaches for color propagation rely
on optical flow computation to propagate colors in videos
either from scribbles or fully colored frames. 
Estimating these correspondence maps is computationally 
expensive and error prone. 
Inaccuracies in optical flow can lead to color artifacts which accumulate over time.
Recently, deep learning 
methods have been proposed to take advantage of semantics for color propagation
in images~\cite{zhang2017real} and videos~\cite{jampani2017video}.
Still, these approaches have some limitations 
and do not yet achieve satisfactory results 
on video content.

\begin{figure*}[t]
	{\renewcommand{\arraystretch}{-1cm}}
	\setlength{\tabcolsep}{0.12em}
	\begin{tabular}{ c | c c c c}
		{\footnotesize Ref. ($k=0$)} 
		& \footnotesize Image PropNet~\cite{zhang2017real} 
		& \footnotesize Style transfer~\cite{li2018closed}
		& \footnotesize SepConv~\cite{niklaus2017video}		
		& \footnotesize Video PropNet~\cite{jampani2017video} \\
		\includegraphics[trim={1cm 0.5cm 1cm 0cm}, clip,width=0.195\linewidth]{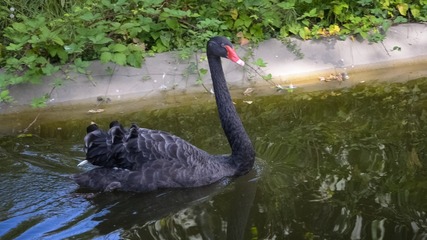} 
		& \includegraphics[trim={0.5cm 0.25cm 1.5cm 0.25cm}, clip,width=0.195\linewidth]{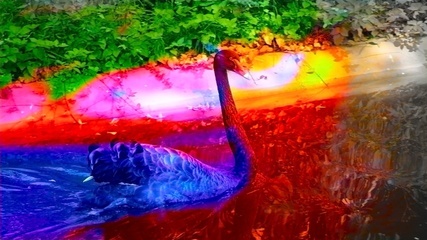}  
		& \includegraphics[trim={0.5cm 0.25cm 1.5cm 0.25cm}, clip,width=0.195\linewidth]{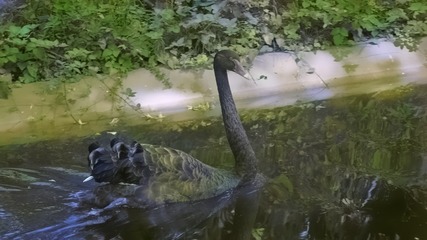} 
		& \includegraphics[trim={0.5cm 0.25cm 1.5cm 0.25cm}, clip,width=0.195\linewidth]{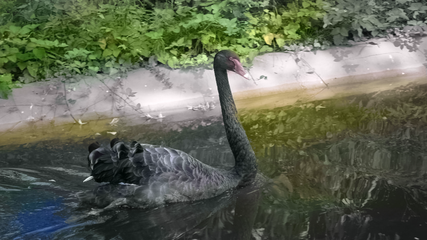}
		&
		\includegraphics[trim={0.5cm 0.25cm 1.5cm 0.25cm}, clip,width=0.195\linewidth]{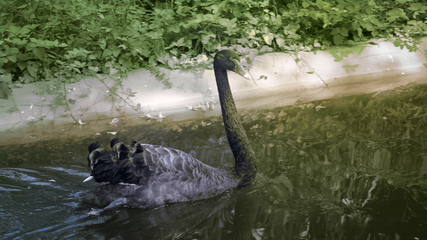} \\
		{\footnotesize Ground Truth ($k=30$)} 
		& \footnotesize Phase-based~\cite{meyer2016phase} 
		 & \footnotesize Bil.Solver~\cite{barron2016fast} 
		 & \footnotesize Flow-based~\cite{xia2016robust} 
		 & \footnotesize Ours \\
		\includegraphics[trim={0.5cm 0.25cm 1.5cm 0.25cm}, clip,width=0.195\linewidth]{results/teaser/gt_30.jpg} 		
		& \includegraphics[trim={0.5cm 0.25cm 1.5cm 0.25cm}, clip,width=0.195\linewidth]{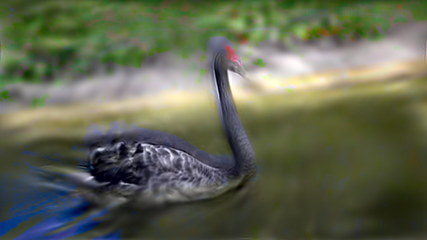}
		& \includegraphics[trim={0.5cm 0.25cm 1.5cm 0.25cm}, clip,width=0.195\linewidth]{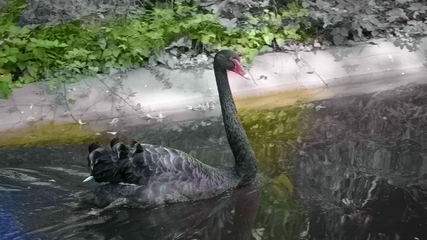}
		&\includegraphics[trim={0.5cm 0.25cm 1.5cm 0.25cm}, clip,width=0.195\linewidth]{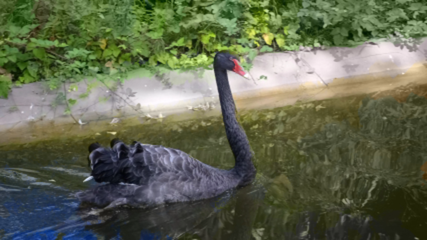}  
		& \includegraphics[trim={0.5cm 0.25cm 1.5cm 0.25cm}, clip,width=0.195\linewidth]{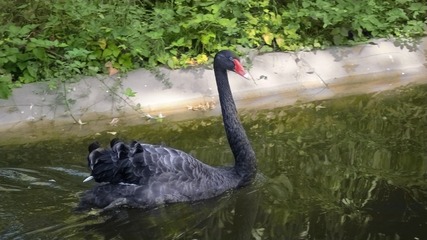} \\		
	\end{tabular}	
	\caption{\textbf{Color propagation after 30 frames ($k=30$).} 
		Our approach is superior to existing strategies 
		for video color propagation. (Image source: \cite{pont2017davis})
		}
	\vspace{-0.4cm}
	\label{fig:teaser}
\end{figure*}

In this work we propose a framework for color propagation in videos
that combines local and global strategies. Given the first frame of a sequence in color, the local strategy warps these 
colors frame by frame based on the motion.
However this local warping becomes less reliable with increasing distance from the 
reference frame. To account for that we propose a global strategy to transfer colors 
of the first frame based on semantics, through deep feature matching. 
These approaches are combined through a fusion and refinement network 
to synthesize the final image. The network is trained on video sequences 
and our evaluation shows the superiority of the proposed method over
image and video propagation methods as well as neural style transfer approaches, see Figure~\ref{fig:teaser}.

Our main contribution is a deep learning architecture, 
that combines local and global strategies for color propagation in videos.
We use a two-stage training procedure 
necessary to fully take advantage of both strategies.
Our approach achieves state-of-the-art results as it is able to maintain better colorization results over a longer time interval compared to a wide range of methods.

\section{Related work}

\subsection{Image and Video Colorization}

A traditional approach to image colorization is to propagate colors or transformation parameters from user scribbles to unknown regions. 
Seminal works in this direction considered low level affinities based 
on spatial and 
intensity distance~\cite{levin2004colorization}.
To reduce user interaction, many directions have been considered such as 
designing better similarities~\cite{luan2007natural}.
Other approaches to improve edit propagation include embedding
learning~\cite{chen2012manifold}, iterative feature
discrimination~\cite{xu2013sparse} or dictionary learning~\cite{chen2014sparse}.
Achieving convincing results for automatic image 
colorization~\cite{cheng2015deep,iizuka2016let}, 
deep convolutional networks have also been
considered for edit propagation~\cite{endo2016deepprop} and interactive
image colorization~\cite{zhang2017real}.
To extend edit propagation to videos, computational efficiency is 
critical and various strategies have been 
investigated~\cite{an2008appprop,yatagawa2014temporally}.


One of the first method considering gray scale video colorization 
was proposed by Welsh~\emph{et al.}~\cite{welsh2002transferring} 
as a frame-to-frame color propagation. 
Later, image patch comparisons~\cite{sykora2004unsupervised}
were used to handle large displacements and rotations. 
However this method targets cartoon content and is not directly adaptable to natural videos.
Yatzi~\emph{et al.}~\cite{yatziv2006fast} 
consider geodesic distance in the 3d spatio-temporal volume to color
pixels in videos and Sheng~\emph{et al.}~\cite{sheng2011colorization} 
replace spatial distance by a distance based on Gabor features.
The notion of reliability and priority~\cite{heu2009image}
for coloring pixels allow better color propagation. 
These notions are extended to entire frames~\cite{xia2016robust},
considering several of them as sources for coloring next gray images.
For increased robustness, Pierre~\emph{et al.}~\cite{pierre2017interactive} 
use a variational model that rely on temporal correspondence maps 
estimated through patch matching and optical flow estimation.

Instead of using pixel correspondences, some recent methods have 
proposed alternative approaches to the video 
colorization problem. Meyer~\emph{et al.}~\cite{meyer2016phase}
transfer image edits as modifications of the phase-based representation
of the pixels. 
The main advantage is that expensive global optimization
is avoided, however propagation is limited to only a few frames. 
Paul~\emph{et al.}~\cite{paul2017spatiotemporal} uses instead of motion vectors the dominant orientations of a 3D steerable pyramid decomposition as guidance for the color propagation of user scribbles.
Jampani~\emph{et al.}~\cite{jampani2017video}, on the other hand, use 
a temporal bilateral network for dense and video adaptive filtering,
followed by a spatial network to refine features.

\subsection{Style Transfer}
Video colorization can be seen as transferring the \emph{color} 
or \emph{style} of the first frame to the rest of the images 
in the sequence.
We only outline the main directions of \emph{color} transfer
as an extensive review of these methods is available 
in~\cite{faridul2016colour}. 
Many methods rely on histogram 
matching~\cite{reinhard2001color}
which can achieve surprisingly good results given their relative 
simplicity but colors could be transferred between incoherent regions.
Taking segmentation into account can help to improve this 
aspect~\cite{tai2005local}. Color transfer between videos
is also possible~\cite{bonneel2013example} by segmenting
the images using luminance and transferring chrominance. 
Recently Arbelot~\emph{et al.}~\cite{arbelot2016automatic}
proposed an edge-aware texture descriptor to guide the colorization. 
Other works focus on more complex transformations such 
as changing the time of the day in photographs~\cite{shih2013data},
artistic edits~\cite{shih2014style} or 
season change~\cite{okura2015unifying}.

Since the seminal work of 
Gatys~\emph{et al.}~\cite{gatys2016image},
various methods based on neural networks have been 
proposed~\cite{li2016combining}.
While most of them focus on painterly results,
several recent works have targeted photo-realistic style 
transfer~\cite{mechrez2017Photorealistic,luan2017deep,li2018closed,he2017neural}.
Mechrez~\emph{et al.}~\cite{mechrez2017Photorealistic} rely on 
Screened Poisson Equation to maintain the fidelity with the style
image while constraining the results to have gradients similar to
the content image. In~\cite{luan2017deep} photo-realism is maintained
by constraining the image transformation to be locally affine in color space.
This is achieved by adding a corresponding loss to the original neural
style transfer formulation~\cite{gatys2015texture}.
To avoid the resulting slow optimization process, patch matching on 
VGG~\cite{he2017neural} features can be used to obtain a guidance image.
Finally, Li~\emph{et al.}~\cite{li2018closed} proposed a two stage
architecture where an initial stylized image, estimated
through whitening and coloring transform (WCT)~\cite{li2017universal}, 
is refined with a smoothing step.

\section{Overview}
\begin{figure}[t]
	\centering
	\includegraphics[width=0.95\columnwidth]{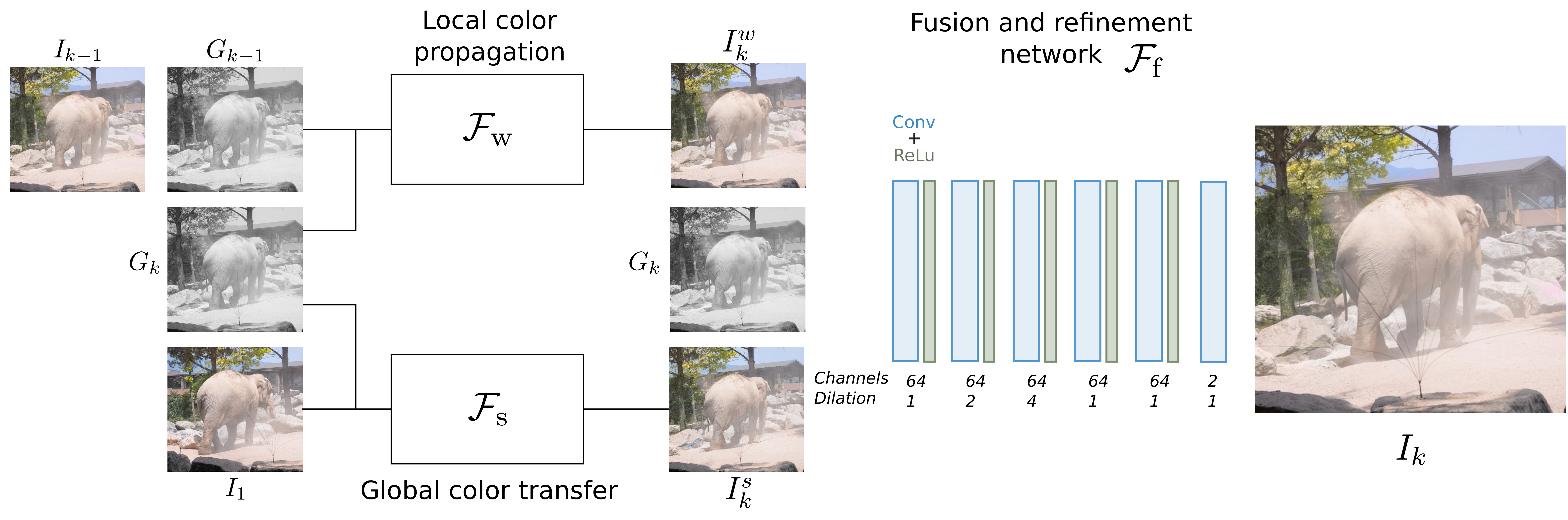}
	\vspace{-0.2cm}
	\caption{\textbf{Overview.} To propagate colors 
		in a video we use both short range and long range color
		propagation. First, the local color propagation network
		$\mathcal{F}_\text{w}$ uses consecutive gray scale frames $G_{k-1}$ and $G_{k}$ 
		to predict spatially adaptive kernels that account for
		motion and re-sampling from $I_{k-1}$. 
		To globally transfer the colors from the reference frame $I_1$ to 
		the entire video a matching based on deep image features is used. 
		The results of these two steps, $I_k^s$
		and $I_k^w$, are together with $G_{k}$ the input to the fusion and refinement network which estimates the final current color frame $I_k$. (Image source:~\cite{pont2017davis})}
	\label{fig:overview}
\end{figure}

The goal of our method is to colorize a gray scale image sequence by propagating the given color of the first frame to the following frames.
Our proposed approach takes into account two complementary aspects: 
short range and long range color propagation, see Figure~\ref{fig:overview}.

The objective of the short range propagation network is to propagate
colors on a frame by frame basis. It takes as input two consecutive 
gray scale frames and estimates a warping function. This warping
function is used to transfer the colors of the previous frame to the next one.
Following recent trends~\cite{xue2016visual,jia2016dynamic,niklaus2017video}, warping is expressed as a convolution process.
In our case we choose to use 
spatially adaptive kernels that account for motion and re-sampling 
simultaneously~\cite{niklaus2017video}, but 
other approaches based on optical flow could be considered as well.

For longer range propagation, simply smoothing warped colors 
according to the gray scale guide image is not sufficient. 
Semantic understanding of the scene is needed to transfer color from the first colored frame of the video to the rest of the video sequence. 
In our case, we find correspondences between pixels of the first frame 
and the rest of the video. Instead of matching pixel colors directly we incorporate semantical information by matching deep features extracted from the frames. 
These correspondences are then used in order to sample colors 
from the first frame.
Besides the advantage for long range color propagation, 
this approach also helps to recover missing colors due to occlusion/dis-occlusion.

To combine the intermediate images of these two parallel stages,
we use a convolutional neural network.
This corresponds to the fusion and refinement stage. 
As a result, the final colored image is estimated by taking advantage 
of information that is present in both intermediate images, 
i.e. local and global color information.


\section{Approach}
Let's consider a grayscale video sequence $\mathcal{G} = \{G_1, G_2, \dots , G_n\}$ of $n$ frames, where the colored image $I_1$ (corresponding to $G_1$)
is available. Our objective is to use the frame $I_1$ to colorize 
the set of grayscale frames $\mathcal{G}$. 
Using a local (frame-by-frame) strategy, colors of $I_1$ can be sequentially
propagated to the entire video using temporal consistency.
With a global strategy, colors present in the first frame $I_1$ can be
simultaneously transfered to all the frames of the video using 
a style transfer like approach. 
In this work we propose a unified solution for video colorization
combining local and global strategies.



\subsection{Local Color Propagation}
\label{local_propagation}
Relying on temporal consistency, our objective is to propagate colors
frame by frame. Using the adaptive convolution approach developed 
for frame interpolation~\cite{niklaus2017video}, one can similarly 
write color propagation as convolution operation on the color image:
given two consecutive grayscale frames $G_{k-1}$ and $G_{k}$, and
the color frame $I_{k-1}$, an estimate of the colored frame $I_{k}$ 
can be expressed as
\begin{equation}
I^\text{w}_k(x,y) = P_{k-1}(x,y) \ast K_{k}(x,y) \;,
\end{equation}
where $P_{k-1}(x,y)$ is the image patch around pixel $I_{k-1}(x,y)$ 
and $K_{k}(x,y)$ is the estimated pixel dependent convolution kernel based on $G_k$ and $G_{k-1}$. 
This kernel is approximated with two 1D-kernels as
\begin{equation}
K_{k}(x,y) = K^v_{k}(x,y) \ast K^h_{k}(x,y) \;.
\end{equation}
The convolutional neural network architecture used to predict these kernels is similar to the one originally proposed for frame 
interpolation~\cite{niklaus2017video},
with the difference that 2 kernels are predicted
(instead of 4 in the interpolation case). Furthermore, we use a softmax layer for kernel prediction which helps to speedup
training~\cite{vogels2018denoising}. 
If we note $\mathcal{F}_w$ the prediction function, 
the local color propagation can be written as
\begin{equation}
I^w_k = \mathcal{F}_{\text{w}}(G_{k}, G_{k-1}, I_{k-1}; \Lambda_{\text{w}}) \;,
\end{equation}
with $\Lambda_{\text{w}}$ being the set of trainable parameters.

\subsection{Global Color Transfer}
\label{global_propagation}
The local propagation strategy becomes less reliable as the frame
to colorize is further away from the first frame. 
This can be due to occlusions/dis-occlusions, new elements appearing 
in the scene or even complete change of background (due to camera panning 
for example). In this case, a global strategy 
with semantic understanding of the scene is necessary. It allows
to transfer color within a longer range both temporally and spatially.
To achieve this, we leverage deep feature extracted with 
convolutional neural networks trained for classification and image segmentation. 
Similar ideas have been developed for style 
transfer and image inpainting~\cite{li2016combining,yang2017high}.

\begin{figure}[t]
	\centering
	\hspace*{-0.5cm} 
	\includegraphics[width=0.95\textwidth]{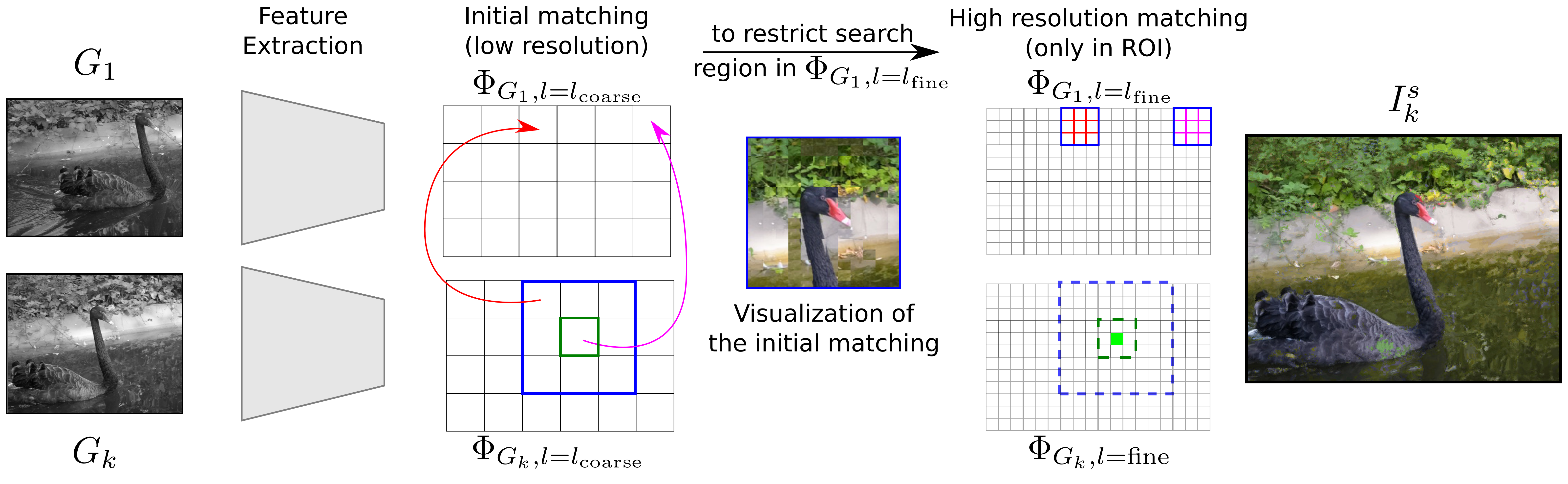} 
	\caption{\textbf{Global Color Transfer.} To transfer the colors of the first 
		frame $I_1$, feature maps $\phi_{G_1}$ and $\phi_{G_k}$
		are extracted from both inputs $G_1$ and $G_k$. 
		First, a matching is estimated at low resolution. 
		This matching performed on features from a deep layer
		($l_\text{coarse}$)	allows to consider more abstract information.
		It is however too coarse to directly copy corresponding 
		image patches. Instead, we use this initial matching to
		restrict the search region when matching pixels using 
		low level image statistics (from level $l_\text{fine}$ feature map).
		Here we show the region of interest (in blue) used 
		to match the pixel in light green. 
		All the pixels sharing the same coarse positions (in dark green rectangle) share the same Region Of Interest (ROI). 
		Using the final matching, $I_1$ colors are transfered to
		the current gray scale image $G_k$. (Image source:~\cite{pont2017davis})}
	\vspace{-0.4cm}		
	\label{fig:style_net}
\end{figure}

Formally, we note $\Phi_{I,l}$ the feature map extracted from the image $I$ 
at layer $l$ of a discriminatively trained deep convolutional neural network.
We can estimate a pixel-wise matching between 
the reference frame $G_1$ and the current frame to colorize $G_k$
using their respective features maps $\Phi_{G_1, l}$ and $\Phi_{G_k, l}$.
Similarity for two positions $\mathbf{x,x'}$ is measured as: 
\begin{equation}
\mathcal{S}_{G_k, G_1}(\mathbf{x},\mathbf{x}^\prime) =
    || \Phi_{G_k, l}(\mathbf{x})~-~\Phi_{G_1, l}(\mathbf{x}^\prime) ||^2_2 \;.
\end{equation}
Transferring the colors using pixel descriptor matching can be written as:

\begin{equation}
\label{eqn:matching}
I^s_k(\mathbf{x}) = I_1(\argmin_{\mathbf{x}^\prime} \mathcal{S}_{G_k, G_1}(\mathbf{x},\mathbf{x}^\prime)) \;.
\end{equation}
To maintain good quality for the matching, while being computationally efficient,
we adopt a two stage coarse-to-fine matching. 
Matching is first estimated for features from a deep layer $l=l_\text{coarse}$. This first matching,
at lower resolution, defines a region of interest for each pixel
in the second matching step of features at level $l=l_\text{fine}$. 
The different levels $l$ of the feature maps correspond 
to different abstraction level. The coarse level matching allows to consider 
regions that  have similar semantics, whereas the fine matching step
considers texture-like statistics that are more effective once a region
of interest has been defined. 
We note $\mathcal{F}_s$ the global color transfer function
\begin{equation}
I^s_k = \mathcal{F}_{\text{s}}(G_{k}, I_1, G_{1}; \Lambda_{\text{s}}) \;,
\end{equation}
with $\Lambda_{\text{s}}$ being the set of trainable parameters.
Figure~\ref{fig:style_net} illustrates all the steps from 
feature extraction to color transfer.
Any neural network trained for image segmentation could be
used to compute the features maps. In our case we use
ResNet-101~\cite{he2016deep}
architecture fine tuned for semantic image segmentation~\cite{chen2018deeplab}. For $l_\text{coarse}$ we use the output of the last layer of the \textit{conv3}-block, while for $l_\text{fine}$ we use the output of the first \textit{conv1}-block (but with stride 1).

\subsection{Fusion and Refinement Network}
\label{fusion_net}
The results we obtain from the local and global stages are complementary.
The local color propagation result is sharp with most 
of the fine details preserved. Colors are mostly well estimated 
except at occlusion/dis-occlusion boundaries where some color bleeding 
can be noticed. 
The result obtained from the global approach is very coarse but
colors can be propagated to a much larger range both temporally 
and spatially. Fusing these two results is learned with a fully 
convolutional neural network. 

For any given gray scale frame $G_k$, the local and global steps 
result in two estimates of the color image $I_k$: $I^w_k$ and $I^s_k$. 
These intermediate results are leveraged by the proposed convolutional 
network (Figure~\ref{fig:overview}) to predict the final output:
\begin{equation}
I_k = \mathcal{F}_{\text{f}}(G_k, I^w_k, I^s_k; \Lambda_{\text{f}}) \;,
\end{equation}
where $\mathcal{F}_{\text{f}}$ notes the prediction function and $\Lambda_{\text{f}}$ the set of trainable parameters.

\vspace{0.2cm}
\noindent\textbf{Architecture details.} The proposed fusion and refinement network consists of 5 convolutional layers with $64$ output channels each followed by a relu-activation function. To keep the full resolution we use strides of $1$ and increase the receptive field by using dilations of $1,2,4,1$ and $1$, respectively. To project the output to the final colors we us another convolutional layer without any activation function. To improve training and the prediction we use instance normalization \cite{ulyanov2016instance} to jointly normalize the input frames. The computed statistics are then also used to renormalize the final output. 

\subsection{Training}
Since all the layers we use are differentiable, the proposed
framework is end-to-end trainable, and can be seen as predicting the
colored frame $\hat{I}_k$ from all the available inputs
\begin{equation}
I_k = \mathcal{F}(G_k, G_{k-1}, I_{k-1}, I_1; \Lambda_{\text{w}}, \Lambda_{\text{s}}, \Lambda_{\text{f}}).
\end{equation}

The network is trained to minimize the total objective function 
$\mathcal{L}$ over the dataset $\mathcal{D}$ consisting of sequences 
of colored and gray scale images. 
\begin{align}
\small{
\Lambda^*_f, \Lambda^*_w = \argmin_{\Lambda_f, \Lambda_w}\mathbb{E}_{I_1, I_2, G_1, G_2 \sim \mathcal{D}} \;\;[\mathcal{L}]
\;.}
\end{align}

\noindent\textbf{Image loss.} We use the $\ell_1$-norm 
of pixel differences which has been shown to lead to sharper results than $\ell_2$~\cite{niklaus2017video,long2016learning,mathieu2015deep}.
This loss is computed on the final image estimate:
\begin{align}
\mathcal{L}_1 = || I_k - \hat{I_k}||_1 \;.
\end{align}

\noindent\textbf{Warp loss.} The local propagation
part of the network has to predict the kernels used
to warp the color image $I_{i-1}$. This is enforced through the
warp loss. It is also computed as the $\ell_1$-norm 
of pixel differences between the ground truth image $I_i$
and $I^w_{i}$:
\begin{align}
\mathcal{L}_w = || I_k - I^w_k||_1 \;.
\end{align}
Since $I^w_k$ is an intermediate result, using more
sophisticated loss functions such as feature loss~\cite{gatys2015texture}
or adversarial loss~\cite{goodfellow2014generative}
is not necessary. All the sharp details will be recovered
by the fusion network.
%
%

\vspace{0.2cm}
\noindent\textbf{Training procedure.} 
To train the network we used pairs of frames
from video sequences obtained from the DAVIS~\cite{perazzi2016benchmark,pont2017davis} dataset 
and Youtube. We randomly extract patches of $256\times256$
from a total of $30$k frames. We trained the fusion net with a batch size of 16
over 12 epochs.

To efficiently train the fusion network we first apply $\mathcal{F}_\text{w}$ and $\mathcal{F}_\text{s}$ separately to all training video sequences.
The resulting images $I^w_k$ and $I^s_k$ 
show the limitations of their respective 
generators $\mathcal{F}_\text{w}$ and $\mathcal{F}_\text{s}$.
The fusion network can then be trained to synthesize the best
color image from these two intermediate results. As input
we provide $G_k$ and the intermediate images $I^w_k$ and $I^s_k$ 
converted to Yuv-color space. 
Using the luminance channel helps the prediction process as it can 
be seen as an indicator on the accuracy of the intermediate results. 
The final image consists of the chrominance values 
estimated by the fusion network and $G_k$ as luminance channel.

\vspace{0.2cm}
\noindent\textbf{Running time.} 
At test time, the matching step is the most computationally involved. 
Still, our naive implementation with TensorFlow computes high 
resolution ($1280\times720$) edit propagation within $5s$
per frame on a Titan X (Pascal).


\section{Results}
For our evaluation we used various types of videos.
This includes videos from DAVIS~\cite{perazzi2016benchmark,pont2017davis},
using the same test set as in~\cite{jampani2017video}, as well as from~\cite{butler2012sintel}. 
We also test our approach on HD videos from the video compression 
dataset~\cite{wang2017videoset}.

\begin{figure*}[t]
	\centering
	\setlength{\tabcolsep}{0.12em}
	\begin{tabular}{ c | p{0.01cm} c c  c c c}
		{\small Reference ($k=0$)} 
		& & & \small Local Only
		& \small Global Only
		& \small Full method
		& {\small Ground Truth}\\
		\includegraphics[trim={7.5cm 0cm 2.5cm 5cm}, clip,width=0.18\linewidth]{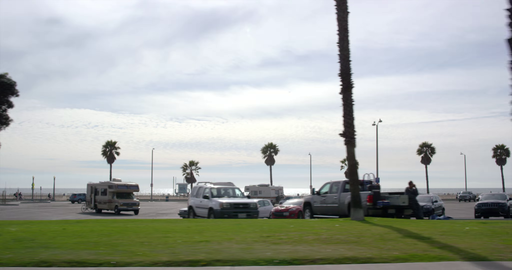}
		& &
		\begin{turn}{90} \small \;\, $k=17$ \end{turn} 
		&
		\includegraphics[trim={7.5cm 0cm 2.5cm 5cm}, clip,width=0.18\linewidth]{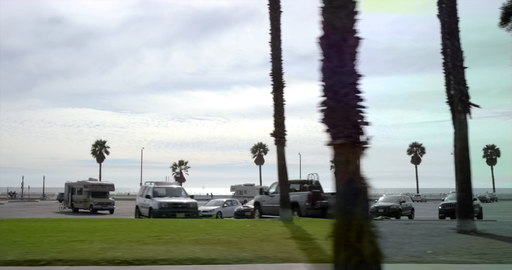}
		&
		\includegraphics[trim={7.5cm 0cm 2.5cm 5cm}, clip,width=0.18\linewidth]{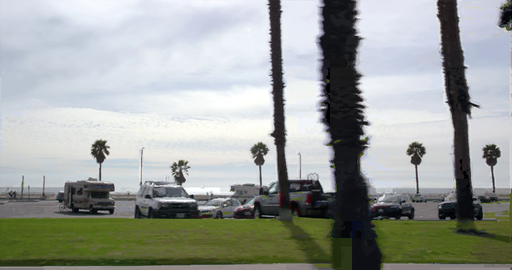}
		& 
		\includegraphics[trim={7.5cm 0cm 2.5cm 5cm}, clip,width=0.18\linewidth]{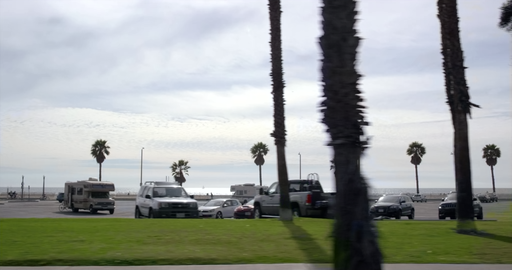}
		& 
		\includegraphics[trim={7.5cm 0cm 2.5cm 5cm}, clip,width=0.18\linewidth]{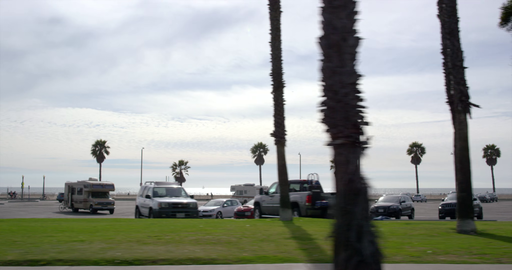}\\
		\includegraphics[trim={3.5cm 0 2cm 0.5cm }, clip,width=0.18\linewidth]{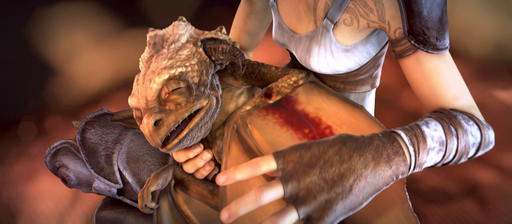}
		& &
		\begin{turn}{90} \small \;\, $k =30$ \end{turn}
		& 
		\includegraphics[trim={3.5cm 0 2cm 0.5cm }, clip,width=0.18\linewidth]{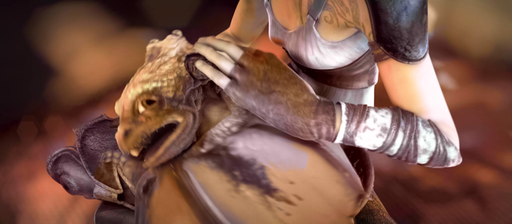}
		&
		\includegraphics[trim={3.5cm 0 2cm 0.5cm }, clip,width=0.18\linewidth]{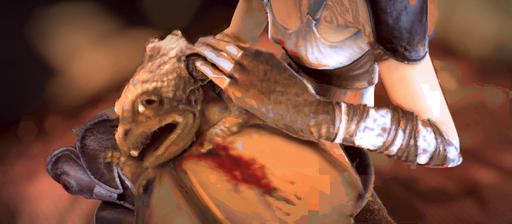}
		& 
		\includegraphics[trim={3.5cm 0 2cm 0.5cm }, clip,width=0.18\linewidth]{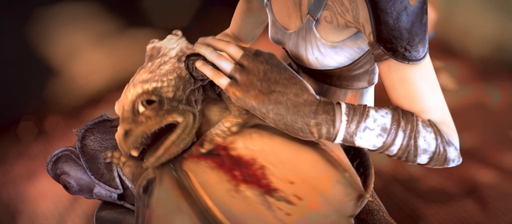}
		& 
		\includegraphics[trim={3.5cm 0 2cm 0.5cm}, clip,width=0.18\linewidth]{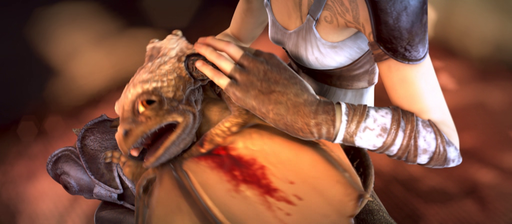}\\
	\end{tabular}	
	\caption{\textbf{Ablation study.} Using local color 
		propagation based on~\cite{niklaus2017video} only preserve details but is 
		sensitive to occlusion/dis-occlusion.
		Using only global color transfer does not 
		preserve details and is not temporally stable.
		Best result is obtained when combining both
		strategies. See Figure~\ref{fig:eval_temporal} for quantitative evaluation. (Image source:~\cite{wang2017videoset,butler2012sintel})}
	\label{fig:ablation}
\end{figure*}

\vspace{0.2cm}
\noindent\textbf{Ablation Study.} To show the importance
of both the local and global strategy, we evaluate both 
configuration. 
The local strategy is more effective for temporal stability 
and details preservation but is sensitive to occlusion/dis-occlusion.
Figure~\ref{fig:ablation} shows an example
where color propagation is not possible due to an occluding object,
and a global strategy is necessary.
Using a global strategy only is not sufficient, as some details
are lost during the matching step and temporal 
stability is not maintained (see video in supplemental material).

\begin{figure*}[t]
	\centering
	\begin{tabular}{ p{0.01cm} c | c c c }
		& {\small
			Ground Truth} & \small Zhang~\emph{et al.}~\cite{zhang2017real} & \small Barron~\emph{et al.}~\cite{barron2016fast} & \small Ours\\
		\begin{turn}{90} \quad~~$k=2$ \end{turn}&
		\adjincludegraphics[trim={0cm 1cm 7.5cm 1cm}, clip,width=0.2\linewidth]{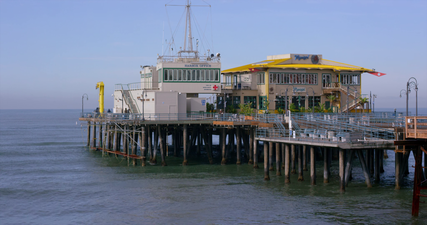} & 
		\adjincludegraphics[trim={0cm 1cm 7.5cm 1cm}, clip,width=0.2\linewidth]{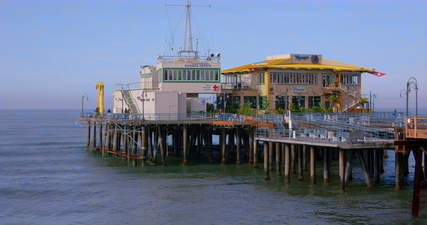} & 
		\adjincludegraphics[trim={0cm 1cm 7.5cm 1cm}, clip,width=0.2\linewidth]{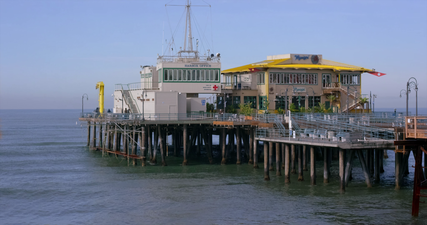} &
		\adjincludegraphics[trim={0cm 1cm 7.5cm 1cm}, clip,width=0.2\linewidth]{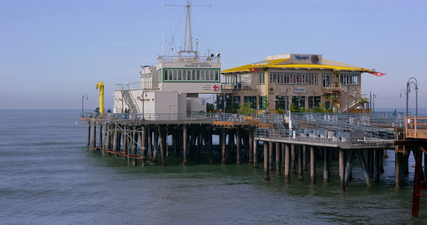} \\
		\begin{turn}{90} \quad~$k=30$ \end{turn}&
		\adjincludegraphics[trim={0cm 0.5cm 7.5cm 1.5cm}, clip,width=0.2\linewidth]{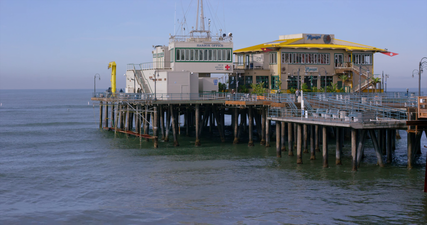} & 
		\adjincludegraphics[trim={0cm 0.5cm 7.5cm 1.5cm}, clip,width=0.2\linewidth]{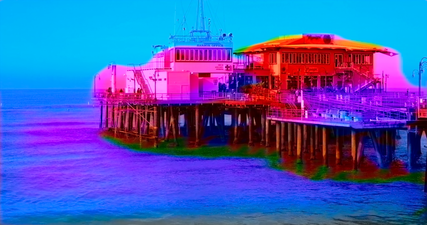} & 
		\adjincludegraphics[trim={0cm 0.5cm 7.5cm 1.5cm}, clip,width=0.2\linewidth]{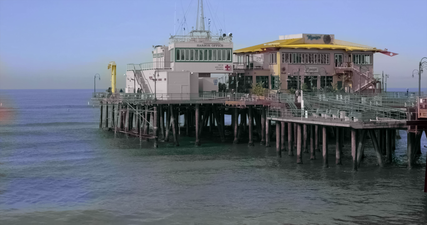} &
		\adjincludegraphics[trim={0cm 0.5cm 7.5cm 1.5cm}, clip,width=0.2\linewidth]{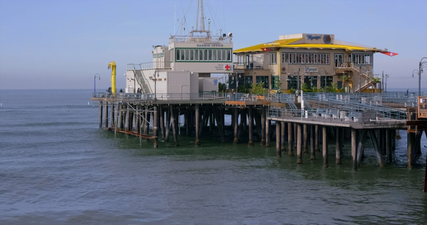} \\
	\end{tabular}	
	\caption{\textbf{Comparison with image color propagation methods.} Methods
		propagating colors in a single image achieve good results on the first frame.
		The quality of the results degrades as the frame to colorize is further
		away from the reference image. (Image source:~\cite{wang2017videoset})}
	\label{fig:results_image_prop}
\end{figure*}

\vspace{0.2cm}
\noindent\textbf{Comparison with image color propagation.} 
Given a partially colored image, 
propagating the colors to the entire image can be achieved using 
the bilateral space~\cite{barron2016fast} or deep learning~\cite{zhang2017real}.
To extend these methods to video, we compute optical flow between consecutive
frames~\cite{zach2007duality} and use it to warp 
the current color image (details provided in supplementary material).
These image based color methods achieve satisfactory color propagation 
on the first few frames (Figure~\ref{fig:results_image_prop}) 
but the quality quickly degrades.
In the case of the bilateral solver, 
there is no single set of parameters 
that performs satisfactorily on all the sequences. 
The deep learning approach~\cite{zhang2017real}
is not designed for videos and drifts towards extreme values.

\begin{figure*}[t]
	{\renewcommand{\arraystretch}{-1cm}}
	\setlength{\tabcolsep}{0.12em}
	\begin{tabular}{ p{0.2cm}  c | c c c c }
		& {\small Ground Truth}
		& \small Phase-based~\cite{meyer2016phase} 
		& \small Video PropNet~\cite{jampani2017video} 
		& \small Flow-based~\cite{xia2016robust}
		& \small Ours \\
		\begin{turn}{90} ~~~~~$k=5$ \end{turn}&
		\adjincludegraphics[trim={7.5cm 2cm {0.05\width} 1cm}, clip,width=0.195\linewidth]{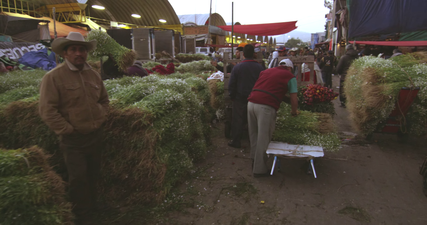} &
		\adjincludegraphics[trim={7.5cm 2cm {0.05\width} 1cm}, clip,width=0.19\linewidth]{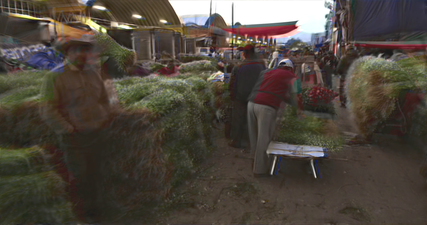} &
		\adjincludegraphics[trim={7.5cm 2cm {0.05\width} 1cm}, clip,width=0.19\linewidth]{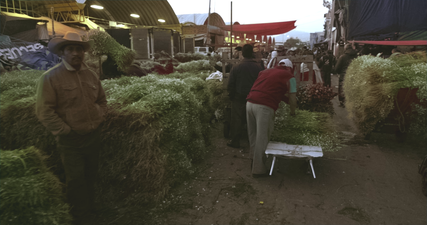} &
		\adjincludegraphics[trim={7.5cm 2cm {0.05\width} 1cm}, clip,width=0.19\linewidth]{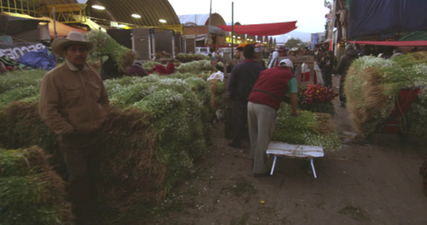} & 
		\adjincludegraphics[trim={7.5cm 2cm {0.05\width} 1cm}, clip,width=0.19\linewidth]{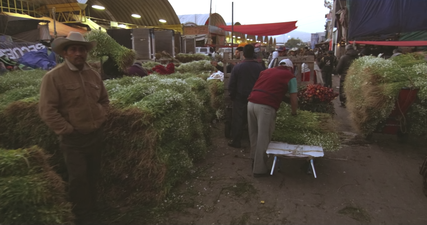}\\
		\begin{turn}{90} ~~~~~$k=25$ \end{turn}&
		\adjincludegraphics[trim={7.5cm 2cm {0.05\width} 1cm}, clip,width=0.19\linewidth]{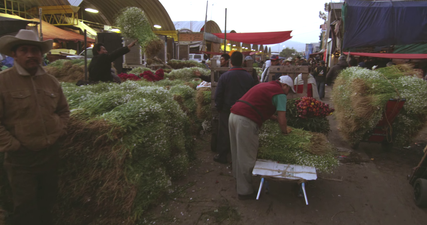} & 
		\adjincludegraphics[trim={7.5cm 2cm {0.05\width} 1cm}, clip,width=0.19\linewidth]{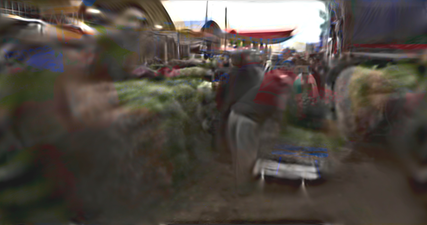} &
		\adjincludegraphics[trim={7.5cm 2cm {0.05\width} 1cm}, clip,width=0.19\linewidth]{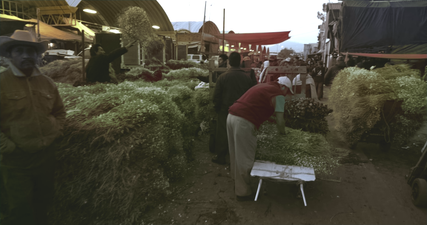} 
		& 
		\adjincludegraphics[trim={7.5cm 2cm {0.05\width} 1cm}, clip,width=0.19\linewidth]{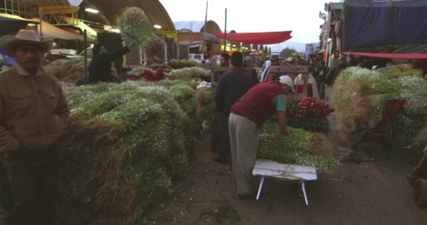} &
		\adjincludegraphics[trim={7.5cm 2cm {0.05\width} 1cm}, clip,width=0.19\linewidth]{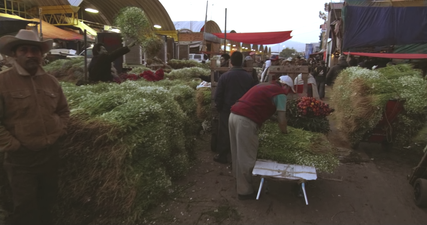} \\
	\end{tabular}	
	\caption{\textbf{Comparison with video color propagation methods.}
	Our approach best retains the sharpness and colors of this video sequence. 
		Our result was obtained in less than one minute while
		the optical flow method~\cite{xia2016robust} 
		needed 5 hours for half the original resolution. (Image source:~\cite{wang2017videoset})
		}
	\label{fig:results_video_prop}
\end{figure*}

\vspace{0.2cm}
\noindent\textbf{Comparison with video color propagation.} 
Relying on optical flow to propagate colors in a video is the most common
approach. In addition to this, Xie~\emph{et al.}~\cite{xia2016robust}
also consider frame re-ordering and use multiple reference frames. 
However, this costly process is limiting as processing $30$ HD frames 
requires several hours. 
Figure~\ref{fig:teaser} and Figure~\ref{fig:results_video_prop} shows that we achieve similar or better
quality in one minute. Phase-based representation can also be used for edit propagation in 
videos~\cite{meyer2016phase}. This original approach to color propagation
is however limited by the difficulty in propagating high
frequencies. 
Recently, video propagation networks~\cite{jampani2017video} were proposed 
to propagate information forward through a video. 
Color propagation is a natural application of such networks.
Contrary to the fast bilateral solver~\cite{barron2016fast}
that only operates on the bilateral grid, 
video propagation networks~\cite{jampani2017video}
benefits from a spatial refinement
module and achieve sharper and better results.
Still, by relying on standard bilateral features (i.e. colors, position, time)
colors can be mixed and propagated from incorrect regions,
which leads to the global impression of washed out colors. 

\begin{figure*}[t]
	\centering	
	\setlength{\tabcolsep}{0.12em}
	\begin{tabular}{ c c | c c c c}
		{\small
			Reference ($k=0$)} & {\small Gray ($k=25$) }& \small \small Li~\emph{et al.}~\cite{li2018closed} & \small Luan~\emph{et al.}~\cite{luan2017deep} & \small Ours\\
		\includegraphics[width=0.195\columnwidth]{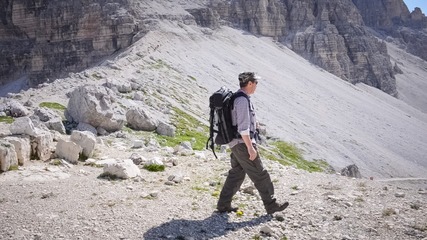}
		& 
		\includegraphics[width=0.195\columnwidth]{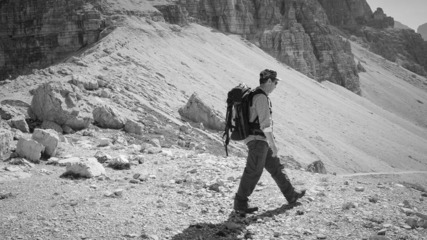}
		& 
		\includegraphics[width=0.195\columnwidth]{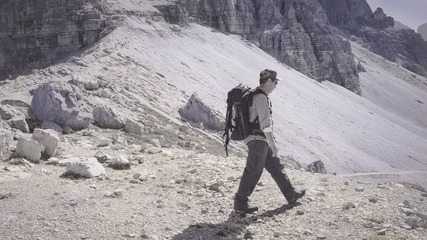}
		&
		\includegraphics[width=0.195\columnwidth]{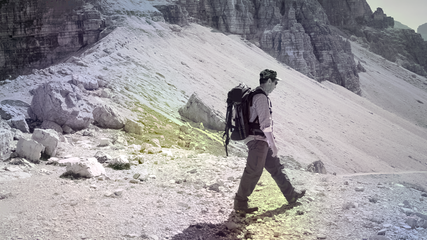}
		& 
		\includegraphics[width=0.195\columnwidth]{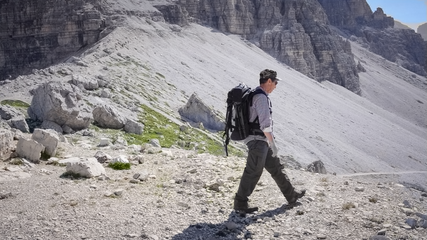}\\
	\end{tabular}	
	\caption{\textbf{Comparison with photo-realistic style transfer.} The reference frame
		is used as style image. (Image source:~\cite{pont2017davis})}
	\label{fig:results_style}
	\vspace{-0.5cm}
\end{figure*}

\vspace{0.2cm}
\noindent\textbf{Comparison with photo-realistic style transfer.} 
Propagating colors of a reference image is the problem solved 
by photo-realistic style transfer methods~\cite{luan2017deep,li2018closed}.
These method replicate the global look but little emphasize 
is put on transferring the exact colors (see Figure~\ref{fig:results_style}). 
\vspace{0.2cm}

\begin{figure}[!h]
	\centering
	\includegraphics[width=0.75\columnwidth]{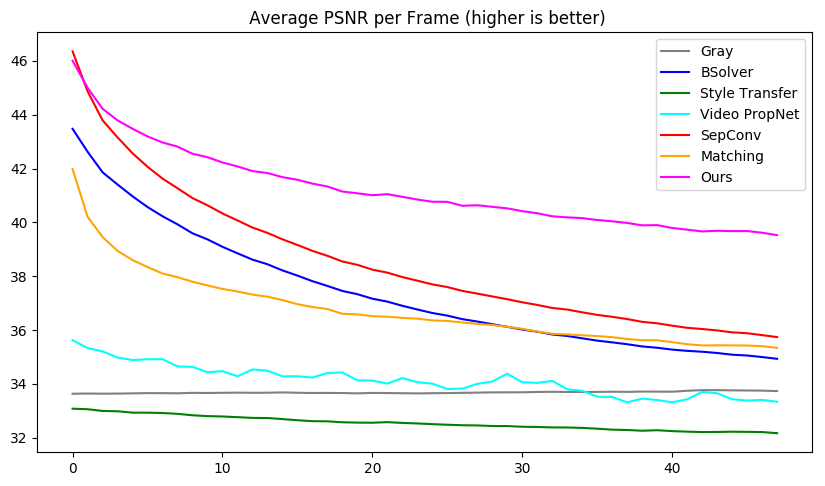} 
\vspace{-0.3cm}
	\caption{\textbf{Temporal evaluation:} 
 The average PSNR error per frame shows the temporal stability
	of our method and its ability to maintain a higher quality 
	over a longer period.}
	\label{fig:eval_temporal}
	\vspace{-0.5cm}
\end{figure}

\noindent\textbf{Quantitative evaluation.}
Our test set consists of $69$ videos which span 
a large range of scenarios with videos containing 
various amounts of motions, occlusions/dis-occlusion, 
change of background and object appearing/disappearing.
Due to their prohibitive running time, some
methods~\cite{xia2016robust,luan2017deep}
are not included in this quantitative evaluation.
Figures~\ref{fig:eval_Nframes} and~\ref{fig:eval_temporal} show the details of this evaluation.
For a better understanding of the temporal behavior of the different methods,
we plot error evolution over time averaged for all sequences.
On the first frames, our results are almost indistinguishable 
from a local strategy (with very similar error values)
but we quickly see the benefit of the global matching strategy.
Our approach consistently outperforms related approaches
for every frame and is able to propagate colors within a much larger 
time frame. Results of the video propagation 
networks~\cite{jampani2017video} vary largely depending 
on the sequence, which explain the inconsistent numerical
performance on our large test set compared to the selected images shown in this paper.

%
%
\begin{figure}[!h]
	\centering
		{\footnotesize
\centering 
\setlength{\tabcolsep}{0.35em}
\renewcommand{\arraystretch}{1.1}
\begin{tabular}{  c | c | c | c | c | c | c | c }

	~~ \textbf{N} ~~ & ~Gray~  & BSolver &  ~Style~ 
	& VideoProp  
	& SepConv \cite{niklaus2017video}
	& Matching 
	& ~~Ours~~ \\ 
	 & & \cite{barron2016fast}  
	 & \cite{li2018closed} 
	&  \cite{jampani2017video}
	&  (local only)
	&  (global only)
	& \\
	\hline 
$10$ & 33.65  & 41.00  & 32.94  & 34.96  & 42.72  & 38.90  & \textbf{43.64}  \\
$20$ & 33.66  & 39.57  & 32.81  & 34.65  & 41.01  & 37.97  & \textbf{42.64}  \\
$30$ & 33.66  & 38.59  & 32.70  & 34.45  & 39.90  & 37.43  & \textbf{42.02}  \\
$40$ & 33.67  & 37.86  & 32.61  & 34.26  & 39.08  & 37.02  & \textbf{41.54}  \\
$50$ & 33.68  & 37.40  & 32.54  & 34.13  & 38.56  & 36.75  & \textbf{41.23}  \\
\end{tabular}
\vspace{0.1cm}
}
	\caption{\textbf{Quantitative evaluation:} 
	Using PSNR in \emph{Lab}-space we compute the average error over the first \textbf{N} frames.}
	\label{fig:eval_Nframes}
\end{figure}

\section{Conclusions}
In this work we have presented a new approach for color
propagation in videos. 
Thanks to the combination of a local strategy, that consists of
a frame by frame image warping, and a global strategy,
based on feature matching and color transfer, 
we have augmented the temporal extent 
to which colors can be propagated.
Our extended comparative results show that the proposed
approach outperforms recent methods in image and video
color propagation as well as style transfer.

\vspace{-0.4cm}
\paragraph{Acknowledgments.}
This work was supported by ETH Research Grant ETH-12 17-1.
\vspace{-0.5cm}

\begin{appendices}
\vspace{-0.1cm}
%
\section{Implementation Details for Comparisons}
\textbf{Image Color Propagation.} 
To extend image color propagation methods ~\cite{barron2016fast,zhang2017real} to video, we compute optical flow between consecutive
frames~\cite{zach2007duality} and use it to warp the current color image
to the next frame.
We compute a confidence measure for the warped 
colors by warping the gray scale image and taking the difference 
in intensities with the original gray frame. 
The warped colors, the confidence maps and the reference gray scale image
can be used to color the second frame using the fast bilateral 
solver~\cite{barron2016fast}.
Using a very conservative threshold, the confidence map 
is binarized to indicate regions 
where colors should be propagated using deep priors~\cite{zhang2017real}.
\end{appendices}

\bibliography{video_arxiv}
\end{document}